*Research Note*

# Narrative Planning: Compilations to Classical Planning

**Patrik Haslum**  PATRIK.HASLUM@ANU.EDU.AU
*Australian National University, Canberra*
*and Optimisation Research Group, NICTA*

## Abstract

A model of story generation recently proposed by Riedl and Young casts it as planning, with the additional condition that story characters behave intentionally. This means that characters have perceivable motivation for the actions they take. I show that this condition can be compiled away (in more ways than one) to produce a classical planning problem that can be solved by an off-the-shelf classical planner, more efficiently than by Riedl and Young's specialised planner.

## 1. Introduction

The classical AI planning model, which assumes that actions are deterministic and that the planner has complete knowledge of and control over the world, is often thought to be too restricted, in that many potential applications problems appear to have requirements that do not fit in this model. Recently, however, it has been shown that some problems thought to go beyond the classical model can nevertheless be solved by classical planners by means of *compilation*, i.e., a systematic remodelling of the problem such that a classical plan for the reformulated problem meets also the non-classical requirements. A striking example is the work of Palacios and Geffner (2006), who showed that conformant planning (generating plans that are robust to certain forms of uncertainty) can be compiled into a classical planning problem. Another example, closer to the topic of this paper, is the work by Porteous, Teutenberg, Pizzi and Cavazza (2011), who use a planner to generate variations of a drama by encoding constraints on the sequencing of events within it.

This paper is about another problem of this kind: planning a *fabula*, meaning the event structure, or "plot", of a story.[1] The fabula planning problem considered was formulated by Riedl and Young (2010). Its main difference from classical planning is a notion of *intentionality*: actions in a story are taken by different characters, and for the story to be considered believable, characters should behave intentionally, i.e., they should have (perceivable) motivations for the actions they take. Riedl and Young argue that "[the fact that classical planners do not take into account character intentions] limits the applicability of off-the-shelf planners as techniques for generating stories," and develop instead a narrative planner, IPOCL, which extends a traditional partial-order causal link planner with a mechanism to enforce that plans respect character intentionality. I will show that fabula planning, as defined by Riedl and Young, can be compiled into a classical planning problem, and hence can in fact be solved by any off-the-shelf classical planner. This does not preclude using an extended formalism, like that introduced by Riedl and Young, which is better suited for the purpose of modelling fabula planning problems, since the compilation is easily automated. The advantage of this is obvious: it allows to bring to bear on the narrative planning problem the entirety of existing, and future, work on algorithms for classical planning, at a dramatically reduced cost in development time and effort. It is not surprising to find that classical planners run on the compiled problem are far more efficient than the IPOCL algorithm, as well as capable of doing more, like finding a set of diverse plans.

---

1. The telling of the story, or *discourse*, is distinct from the fabula (e.g., Gervás 2009). The aforementioned work by Porteous et al. can be seen as the application of planning to generating different discourses of a given fabula.

There are different theories about what distinguishes a story from an arbitrary sequence of events, i.e., what gives it its "storiness" (e.g., Gervás 2009; Mateas & Sengers, 1999). My aim is not to criticise the particular model of narrative planning proposed by Riedl and Young but merely to show that the criterion they adopted – character intentionality – can be achieved by a classical planner without modification, by simply restating the problem that is given to the planner to solve. Whether this can be done also for other models of narrative generation is an open question.

## 2. Narrative Planning and Intentional Plans

*"This is a story about how King Jafar becomes married to Jasmine. There is a magic genie. This is also a story about how the genie dies."*
               (From the textual representation of the story generated by IPOCL; Riedl & Young 2010.)

Riedl and Young observe that "there are many parallels between plans and narrative at the level of fabula." Both are sequences of events that change the state of the (story) world. For a story to be perceived as coherent and plausible, the event sequence must be logically possible (i.e., preconditions achieved before an event takes place) and connected by causes and effects (i.e., each event contributes something to the story, by setting the stage for later events). The goal of a story, in this view, is the end state that the story's author has in mind; Riedl and Young call this the *story outcome*, to distinguish it from character goals. For a story to be believable, characters in it should appear to be *intentional agents*: a character's actions should not only be possible, and contribute to the outcome of the story, but should be perceivable as contributing to the goals that the character has (which are not necessarily the same as the author's goal). This can be seen as a non-redundancy requirement on subsets of the plan: each action done by each character in the story should directly or indirectly contribute to achieving a goal of the character. Of course, goals of a character can change throughout the course of the story, as they are influenced by other characters, or events in the world around them. But each such change of a character's goals must also have a cause.

To illustrate narrative planning, and to evaluate the believability of plans generated by IPOCL, Riedl and Young (appendix A.1, p. 254–256) use the following small example scenario. The dramatis personae are:
- King Jafar, who lives in The Castle;
- Aladdin, a Knight loyal to Jafar;
- Jasmine, a beautiful woman, who also lives in The Castle;
- a Genie, who is imprisoned in a Magic Lamp; and
- a Dragon, who lives at The Mountain and possesses the Lamp at the start of the story.

Characters can travel between the two locations. A knight can slay a monster (only the Dragon and the Genie are monsters). A character can take things from a dead character ("pillage"), and can give things to another character (the Lamp is the only item of interest). A character who has the Lamp can summon the Genie, thereby gaining control over it. The Genie, by magic, can cause a character to fall in love with another character. Two characters who are in love, and not otherwise engaged, can marry. The goals of the story are (married Jafar Jasmine) and (dead Genie). Note that these goals represent the story outcome; they are not (initially) intended by any character.

Riedl and Young distinguish two types of planning actions: *intentional actions*, which correspond to actions taken by one or more story characters (*actors* of the action), and *happenings*, which do not have actors and correspond to accidental events, forces of nature, etc. The classifications of actions as intentional or happenings, and the assignment of the role of actor(s) to parameters of intentional actions, are part of the domain theory. Examples of happenings in the scenario above are for a character to fall in love with another who is beautiful, and for a scary monster to frighten another character.

Character intentions are modelled by modal literals of the form (intends $A$ $f$), where $A$ is a character and $f$ is a fact, i.e., a normal literal. Intentions arise as an effect of actions, either happenings or character actions. For example, the happening (fall-in-love ?man ?woman) has the effect (loves ?man ?woman) and establishes the intention (intends ?man (married ?man ?woman)). Similarly, the action (deliver-witty-insult ?speaker ?hearer ?victim), in which ?speaker is the actor, could have the effect (amused ?hearer), but also the unintended (by the speaker) effect (intends ?victim (dead ?speaker)). A special category of actions that cause intentions are "delegating" actions, where one character commands (or persuades, or bribes, or otherwise influences) another to achieve something. For example, (order ?king ?knight ?goal) has the effect (intends ?knight ?goal).

Riedl and Young define their notion of intentionality in the context of partially ordered causal link (POCL) plans.[2] The following definition summarises definitions 3, 5 and 6 (pp. 232–234) in their article:

**Definition 1** *An* intentional plan *is one in which every occurrence of an intentional action is part of some frame of commitment. A* frame of commitment *is a subset $S'$ of steps (i.e., action occurrences) in the plan, associated with a modal literal (intends $A$ $g$), satisfying four requirements: (1) Character $A$ is an actor of every step in $S'$. (2) There is a final step $s_{fin} \in S'$ that makes $g$ true. (3) There is a motivating step $s_m$ in the plan, which adds (intends $A$ $g$) and which precedes all steps in $S'$. We'll say there is a motivational link from $s_m$ to every step in the frame of commitment, $S'$. Note that $s_m$ is not part of $S'$. (4) From each step in $S'$ other than $s_{fin}$ there is a path of causal or motivational links to $s_{fin}$. A* complete (fabula) plan *is one that is both intentional and valid in the classical sense.*

Condition (4) above departs slightly from the definitions stated by Riedl and Young: they require only that each step in $S'$ temporally precedes $s_{fin}$. That would appear to be too promiscuous, since it allows any unrelated action to be incorporated into a frame of commitment by adding spurious temporal constraints. Their IPOCL algorithm, however, will only incorporate a step into an existing frame of commitment if the step has a causal link to a step already in the frame, or can serve as motivating step for a frame of commitment whose final step has a causal link to a step already in the frame (cf. items 1 and 2 on page 235 of their article). Hence, the frames that their algorithm generates always satisfy condition (4).

## 3. Compilation 1: Explicit Justification Tracking

The first compilation is based on explicit tracking of the justifications, in the form of causal and motivating links, for actions in the plan. It is inspired by the work of Karpas and Domshlak (2011) on pruning redundant action sequences from the search space, which also relies on a notion of justification of actions.

---

2. A detailed account of POCL planning can be found in the paper by McAllester and Rosenblitt (1991) and in most AI textbooks.

We will use three kinds of modal literals: (intends $A$ $f$) and (delegated $A$ $f$), where $A$ is a character and $f$ a fact, and (justified $f$ $I$), where $f$ is a fact and $I$ an intention, i.e., a modal literal of the first form. The intends modality is part of the narrative planning problem specification, where it can appear in action effects and in the initial state. The other modalities are used only to describe the compilation. Of course, modal conditions cannot be expressed directly in a classical planning formalism like PDDL. In a PDDL model, they are replaced by a separate "modal predicate" for each predicate (resp. combination of two predicates) that can appear in a non-modal fact, whose arguments is the concatenation of all arguments in the modal literal. For example, (intends Aladdin (has Jafar Lamp)) is replaced by (intends-has Aladdin Jafar Lamp), and (justified (at Aladdin Mountain) (intends Aladdin (has Jafar Lamp))) is replaced by (justified-at-has Mountain Aladdin Aladdin Jafar Lamp).

In the compiled problem, each intentional action is associated with an intention of the action's actor(s). This intention is a precondition of the action. If the action itself does not achieve the intention, it creates an outstanding obligation to make use of at least one of its effects to achieve the precondition of some other action, done by the same actor, that contributes, directly or indirectly, to achieving the intention. This is modelled by the justified modality. As explained earlier, actions can have modal effects of the form (intends $B$ $f$), i.e., to make a character (different from the actor) intend a goal. This is modelled by the delegated modality.

For each intentional action, $(a\ \vec{x})$, of the narrative planning problem, the compiled problem has one distinct action for each combination of an intention (intends $x_A$ $(p\ \vec{y})$), where $x_A$ is the parameter that represents the actor of $(a\ \vec{x})$, and an effect $(e\ \vec{z})$ of $(a\ \vec{x})$. We name this compiled action ($a$-$e$-because-intends-$p$ $\vec{x}$ $\vec{y}$), and call $(e\ \vec{z})$ the *chosen effect*. Note that the parameters $\vec{z}$ of the chosen effect are composed from a subset of the parameters $\vec{x}$ of the action, and possibly explicit constants. Action ($a$-$e$-because-intends-$p$ $\vec{x}$ $\vec{y}$) can be read as "character $x_A$ performs action $(a\ \vec{x})$ to achieve the effect $(e\ \vec{z})$ as a step towards achieving the character's intended goal $(p\ \vec{y})$". If $(e\ \vec{z})$ can unify with $(p\ \vec{y})$, the action must be further broken into two cases: one where they are forced to equal and one where they are forced to be distinct. For an intentional action with more than one actor, the compiled problem must have a distinct action for each (possible and relevant) choice of effect and intention for each actor. For a happening (i.e., action without an actor) there is only one corresponding action in the compiled problem.

The justified and delegated modalities combine to track causal and motivational links in the compiled problem. All (possible and relevant) justified literals are true in the initial state, and required to be true in the goal state. Action ($a$-$e$-because-intends-$p$ $\vec{x}$ $\vec{y}$) makes the chosen effect *unjustified*, by deleting (justified $(e\ \vec{z})$ (intends $x_A$ $(p\ \vec{y})$)). Since the goal requires all justified literals to hold, the plan must include some action, by the same actor and with the same intention, whose precondition requires $(e\ \vec{z})$; no other action will make (justified $(e\ \vec{z})$ (intends $x_A$ $(p\ \vec{y})$)) true again. If the chosen effect is a modal literal (intends $z_A$ $(q\ \vec{z'})$) it is the subgoal $(q\ \vec{z'})$ that becomes unjustified, and it also becomes "delegated" to the second character, $z_A$. This provides the motivational link from the action ($a$-$e$-because-intends-$p$ $\vec{x}$ $\vec{y}$) to any action that $z_A$ takes to achieve $(q\ \vec{z'})$. Delegation ends when the character achieves the goal. While a goal is delegated, no other character may achieve the goal. This ensures the step that created the delegation is eventually justified, by the character who performed it making use of the achieved fact $(q\ \vec{z'})$.

Let $(a\ \vec{x})$ be an intentional action, $x_A$ the parameter that represents its actor, $(e\ \vec{z})$ the chosen effect and (intends $x_A$ $(p\ \vec{y})$) the intention of the actor. The preconditions of the compiled action ($a$-$e$-because-intends-$p$ $\vec{x}$ $\vec{y}$) are:

(1) all preconditions of $(a\ \vec{x})$;

(2) (intends $x_A$ $(p\ \vec{y})$);

(3a) $\neg \exists w$ (delegated $w$ $(q\ \vec{z'})$), for each effect of $(a\ \vec{x})$ that is of the form (intends $z_A$ $(q\ \vec{z'})$);

(3b) $\neg \exists w \neq x_A$ (delegated $w$ $(p\ \vec{y})$), if $(p\ \vec{y})$ is an effect of $(a\ \vec{x})$;

(3c) $\neg \exists w$ (delegated $w$ $(q\ \vec{z'})$), for any other effect $(q\ \vec{z'})$ of $(a\ \vec{x})$ that is not an intends modal literal.

(4) $\neg$(intends $z_A$ $(q\ \vec{z'})$), if $(e\ \vec{z})$ is a modal literal of the form (intends $z_A$ $(q\ \vec{z'})$).

In a plan for the compiled problem, sets of actions with the same associated intention form a frame of commitment. Precondition (2) ensures all steps in that frame are preceded by a motivating step. Precondition (4) ensures there is at most one (intentional) motivating step. Preconditions (3a–c) ensure that no action can be taken that delegates (a) or achieves (b–c) a goal that is already delegated to another character.

The effects of the compiled action are:

(1) all effects of $(a\ \vec{x})$;

(2) (justified $(q\ \vec{v})$ (intends $x_A$ $(p\ \vec{y})$)), for each (non-static) precondition $(q\ \vec{v})$ of $(a\ \vec{x})$.

(3a) $\neg$(justified $(q\ \vec{z'})$ (intends $x_A$ $(p\ \vec{y})$)) and (delegated $z_A$ $(q\ \vec{z'})$), if $(e\ \vec{z})$ is a modal literal of the form (intends $z_A$ $(q\ \vec{z'})$);

(3b) $\neg$(justified $(e\ \vec{z})$ (intends $x_A$ $(p\ \vec{y})$)), if $(e\ \vec{z})$ is not an intends literal and $(e\ \vec{z})$ does not equal $(p\ \vec{y})$;

(4) $\neg$(delegated $x_A$ $(e\ \vec{z})$), if $(e\ \vec{z})$ equals $(p\ \vec{y})$;

Effects (2) and (3) make the preconditions of the action justified, and the chosen effect unjustified, as explained above. If the chosen effect is a modal intends literal (case 3a), it is the intended subgoal that becomes unjustified, and also delegated to the other character. Effect (4) ends the delegation of a goal when the action is the final step in a frame of commitment. (Note, however, that the action will have this effect even if the goal was not delegated; this does not matter.)

If the chosen effect $(e\ \vec{z})$ can be unified with $(p\ \vec{y})$, the compiled action must be split into two: one with the additional precondition $\vec{z} = \vec{y}$, ensuring that they are equal, and one with the additional precondition $\vec{z} \neq \vec{y}$, ensuring that they are not. This is necessary since the effects of the compiled action depend on whether $(e\ \vec{z})$ equals $(p\ \vec{y})$ or not. Furthermore, as mentioned above, if the original action $(a\ \vec{x})$ has more than one actor, the compiled problem has one action for every combination of an intention and a chosen effect for each actor. In this case, conditions on $(p\ \vec{y})$ and $(e\ \vec{z})$ in the schema above should be interpreted for each actor separately. That is, if $(p^i\ \vec{y}^i)$ and $(e^i\ \vec{z}^i)$ are the intention and chosen effect of actor $x_A^i$, the compiled action has the effect $\neg$(justified $(e^i\ \vec{z}^i)$ (intends $x_A^i$ $(p^i\ \vec{y}^i)$)) if $(e^i\ \vec{z}^i)$ is not an intends literal and $(e^i\ \vec{z}^i)$ does not equal $(p^i\ \vec{y}^i)$ (item 3b), regardless of whether $(e^i\ \vec{z}^i)$ equals $(p^j\ \vec{y}^j)$ for some $j \neq i$, or vice versa.

To illustrate the compilation, consider the following action from the example scenario by Riedl and Young (appendix A.1, p. 255), here written in a more PDDL-like syntax:

```
(:action slay
  :parameters (?knight - knight ?monster - monster ?where - place)
  :actors (?knight)
  :precondition (and (alive ?knight) (at ?knight ?where) (alive ?monster) (at ?monster ?where))
  :effect (and (not (alive ?monster)) (dead ?monster)))
```

The actor of this action is the knight. Consider the intention (intends ?knight (dead ?who)). The action has only one relevant choice of effect, (dead ?monster) (the negative literal does not appear in any action precondition or the goal). However, since the intention unifies with the chosen effect, the compiled problem must still include two actions, one for ?who = ?monster and one for ?who ≠ ?monster:

```
(:action slay-1-because-intends-dead
  :parameters (?knight - knight ?monster - monster ?where - place)
  :precondition (and (alive ?knight) (at ?knight ?where) (alive ?monster) (at ?monster ?where)
                (intends ?knight (dead ?monster))
                (not (exists (?c) (and (not (= ?c ?knight)) (delegated ?c (dead ?monster))))))
  :effect (and (not (alive ?monster)) (dead ?monster)
           (justified (at ?knight ?where) (intends ?knight (dead ?monster)))
           (justified (at ?monster ?where) (intends ?knight (dead ?monster)))
           (not (delegated ?knight (dead ?monster)))))

(:action slay-2-because-intends-dead
  :parameters (?knight - knight ?monster - monster ?where - place ?who - monster)
  :precondition (and (alive ?knight) (at ?knight ?where) (alive ?monster) (at ?monster ?where)
                (intends ?knight (dead ?who))
                (not (exists (?c) (delegated ?c (dead ?monster))))
                (not (= ?who ?monster)))
  :effect (and (not (alive ?monster)) (dead ?monster)
           (justified (at ?knight ?where) (intends ?knight (dead ?who)))
           (justified (at ?monster ?where) (intends ?knight (dead ?who)))
           (not (justified (dead ?monster) (intends ?knight (dead ?who))))))
```

(The alive literals don't need justification, because there is no way to make them true unless true initially.) Corresponding to the intention (intends ?knight (has ?who ?what)) the compiled problem will have the action:

```
(:action slay-because-intends-has
  :parameters (?knight ?monster ?where ?who ?what)
  :precondition (and (alive ?knight) (at ?knight ?where) (alive ?monster) (at ?monster ?where)
                (intends ?knight (has ?who ?what))
                (not (exists (?c) (delegated ?c (dead ?monster)))))
  :effect (and (not (alive ?monster)) (dead ?monster)
           (justified (at ?knight ?where) (intends ?knight (has ?who ?what)))
           (justified (at ?monster ?where) (intends ?knight (has ?who ?what)))
           (not (justified (dead ?monster) (intends ?knight (has ?who ?what))))))
```

To prove the correctness of the compilation in general, we will need the concept of a *toggling* action (Hickmott & Sardina, 2009). An action is toggling w.r.t. an effect of the action iff the action's precondition implies the negation of the effect. That is, if the action makes true a fact $f$, its precondition must include $\neg f$, or some fact $f'$ that is mutex with $f$, and if the action makes $f$ false, it must require $f$ to be true. An action that is not toggling can be transformed into an equivalent set of actions that are, though the size of this set is exponential in the number of non-toggling effects of the original action.

**Theorem 2** *Let $P$ be a narrative planning problem, in which each action is toggling w.r.t. its effects. Let $P'$ be the compiled problem as described above. Every plan $S$ for $P'$ is intentional.*

**Proof:** Consider a step $s$ in $S$, that is an instance of an action ($a$-$e$-because-intends-$p$ $\vec{x}$ $\vec{y}$). Let $A$ be the actor (i.e., the constant bound to the actor parameter $x_A$ of $a$) and ($e$ $\vec{z}$) the chosen effect of ($a$ $\vec{x}$). The action will be part of a frame of commitment with the goal ($p$ $\vec{y}$). By construction, the precondition of (a-e-because-intends-$p$ $\vec{x}$ $\vec{y}$) includes (intends $A$ ($p$ $\vec{y}$)). There must be a motivating step that establishes this precondition, as otherwise $S$ would not be classically valid.

If ($e$ $\vec{z}$) equals ($p$ $\vec{y}$), then $s$ is itself the final step in the frame of commitment.

If ($e$ $\vec{z}$) does not equal ($p$ $\vec{y}$) and is not a modal literal, then (a-e-because-intends-$p$ $\vec{x}$ $\vec{y}$) destroys (justified ($e$ $\vec{z}$) (intends $A$ ($p$ $\vec{y}$)). Since all such literals are goals in $P'$, there must be a later step, $s'$, that re-establishes it. By construction, this can only be an action that has $A$ as actor, (intends $A$ ($p$ $\vec{y}$)) as the associated intention, and ($e$ $\vec{z}$) as a precondition. If there is no step between $s$ and $s'$ that adds ($e$ $\vec{z}$), there must be a causal link labelled with ($e$ $\vec{z}$) from $s$ to $s'$ (since actions are toggling, ($e$ $\vec{z}$) was not true before $s$). There cannot be only a step $s_{\text{add}}$ between $s$ and $s'$ that adds ($e$ $\vec{z}$), because if so, the action associated with $s_{\text{add}}$ would be applied in a state where one of its effects, ($e$ $\vec{z}$), is already true, and thus not toggling. Suppose there are steps $s_{\text{del}}$ and $s_{\text{add}}$ taking place between $s$ and $s'$, such that $s_{\text{del}}$ destroys ($e$ $\vec{z}$) and $s_{\text{add}}$ makes it true again; if there are several such steps, let $s_{\text{del}}$ be the first and $s_{\text{add}}$ the last, so that there is a causal link from $s_{\text{add}}$ to $s'$. Since actions are toggling $s_{\text{del}}$ requires ($e$ $\vec{z}$), so there is a causal link from $s$ to $s_{\text{del}}$. If there is no chain of causal links from $s_{\text{del}}$ to $s_{\text{add}}$, the subplan consisting of steps up to and including $s$ and all causal predecessors of $s_{\text{add}}$ (which do not include $s_{\text{del}}$) must be executable, and results in an execution where the action associated with $s_{\text{add}}$ is again applied in a state where one of its effects, ($e$ $\vec{z}$), is already true, and hence is not toggling. Thus, there must be a chain of causal links from $s_{\text{del}}$ to $s_{\text{add}}$, and therefore from $s$ to $s'$. Since $s'$ is part of the same frame of commitment as $s$ (it has the same motivating intention), and there can only be a finite number of steps in this frame of commitment that causally follow $s$, repeated application of this reasoning leads to the conclusion that there must be a chain of causal links from $s$ to the final step of the frame.

If ($e$ $\vec{z}$) is a modal literal of the form (intends $z_A$ ($q$ $\vec{z'}$)), (a-e-because-intends-$p$ $\vec{x}$ $\vec{y}$) destroys (justified ($q$ $\vec{z'}$) (intends $A$ ($p$ $\vec{y}$)), and adds (delegated $z_A$ ($q$ $\vec{z'}$)). As above, there must be a step $s'$, in the same frame of commitment as $s$, that re-establishes (justified ($q$ $\vec{z'}$) (intends $A$ ($p$ $\vec{y}$)). By construction, (delegated $w$ ($q$ $\vec{z'}$)) can be true for at most one character $w$ at any time (any action that adds a delegation requires that no other character has it), and only the character currently holding the delegation of ($q$ $\vec{z'}$) can make it true. Thus, there is (by the same argument as above) a causal chain from the final step of the delegates frame of commitment with goal ($q$ $\vec{z'}$) to $s'$. Because actions in the compiled problem are toggling w.r.t. intends literals, step $s$ must have a causal chain to the precondition (intends $z_A$ ($q$ $\vec{z'}$)) of each action in the frame of commitment of the delegate, and thus serves as the motivating step for this frame. Thus, there is a chain of motivating and causal links from $s$ to $s'$, and following the same argument as above, therefore from $s$ to the final step of the frame of commitment that $s$ belongs to. □

It may be noted that some apparently reasonable story plans are disallowed. For example, a character cannot delegate a goal that he himself intends to another character. This, however, is a consequence of Riedl and Young's definition of intentional plans, not of the compilation (and hence applies also to the IPOCL planner): the final step in a frame of commitment must achieve the intended goal and must be an action by the actor that holds this intention (conditions 1 & 2 of Definition 1). This rules out delegating one's own goals. If desired, it would not be difficult to modify the compilation to allow this kind of secondary delegation: it requries only adding the exception $w \neq x_A$ to precondition (3a) and an effect like (4) for this case. A plan also cannot have a character trying and failing by multiple means to achieve his goals. Again, this is a consequence of Riedl and Young's definition, not of the compilation: every action taken by a character must have a chain of causal or motivational links to the final step (condition 4 of Definition 1). This rules out characters taking actions that prove ultimately futile.

Theorem 2 shows that the compilation is sound. The question of whether it is also complete, i.e., whether existence of an intentional plan for a narrative planning problem $P$ always implies existence of a plan for the compiled problem $P'$, is somewhat complicated. At first glance, given an intentional plan $S$ for $P$, it appears a plan $S'$ for the compiled problem $P'$ could be constructed by selecting for each action $a$ in $S$ a suitable representative $a$-...-because-..., with intentions and chosen effects to match the frames of commitment to which $a$ belongs in $S$. Since $S$ is intentional, each frame of commitment is preceded by a motivating step, ensuring the intends preconditions of the compiled action are satisifed, and if $a$ is not the final step, there is a causal link from at least one of its effects to another step, ensuring that deleted justified literals are restored. There is, however, one point where the correspondence can fail, due to the restriction of the compiled problem that a delegated goal can only be achieved by the character that it was delegated to: Suppose character $A$ delegates goal $g$ to character $B$, i.e., character $A$ performs an intentional action whose only (relevant) effect is (intends $B$ $g$). For the plan to be intentional, there must be a frame of commitment belonging to character $B$, with the associated intention (intends $B$ $g$); this frame must have a final step which achieves $g$, and that step must be the source of a causal link to some step performed by character $A$, belonging to same the frame of commitment of $A$ as the step that established the motivation. Yet, nothing prevents another character, $C$, from achieving $g$ for his own purposes, as long as character $B$ also achieves $g$. The compiled problem, however, does not allow character $C$ to achieve $g$ as long as it is delegated to $B$, i.e., in between the motivating step by $A$ and the final step by $B$. This could be remedied through a more elaborate justification tracking mechanism, that distinguishes the same fact when achieved by different characters.

From a practical perspective, the combinations of actions with intentions, and modal literals, present in the compiled problem can be restricted to those that are "possible and relevant". For example, the initial state and goal only needs to include those justified literals that can actually be negated by some possibly applicable action (which can be found by standard relaxed reachability analysis). In the example scenario, the fact that a character has the Lamp can never causally contribute to, e.g., the goal of the character having another item (there are no other items to have) or the goal of murdering another character. Thus, actions like pillage-because-intends-dead or order-has-because-intends-has and order-has-because-intends-dead can never be part of valid plan. Most of this information could be found by simple techniques like back-chaining relevance analysis.

Applying the compilation to Riedl and Young's example scenario, and applying a classical planner, using forward-chaining A* search with the LM-Cut heuristic (Helmert & Domshlak, 2009), to the compiled problem, produces the plan shown in figure 1. The planner outputs a sequence of actions, which is transformed to a partially ordered plan by a polynomial time post-processing step (Bäckström, 1998). Enumerating all shortest plans reveals two variations: one in which Jafar travels back to the Castle to marry Jasmine, and one in which Jafar orders Aladdin to bring him the Lamp, and both climactic events (the wedding and Aladding slaying the Genie) take place at the Castle. (The latter is the one Riedl and Young report was found by IPOCL, shown in Figure 15, p. 259, of their article.) Note that it is not possible for Jafar to command Aladdin to make (loves Jasmine Jafar) true, because Aladdin has no means to achieve this goal other than by delegating it to the Genie, which, as explained above, is not permitted by the definition of a frame of commitment.

Finding shortest plans is not an end in itself: rather, it is a side effect of the fact that a planner will usually seek to achieve the story outcome in the simplest way. This can be somewhat at odds with making the story "interesting". Porteous and Cavazza (2009) argue that *complexification*, i.e., making the story more convoluted in order to make it more interesting, can be achieved by posting additional author goals in the form of PDDL3 trajectory constraints, specifying that some fact must be achieved at some point in the plan; that some fact must never be true at any point; or that some fact must be achieved before another. PDDL3 trajectory constraints can also be compiled away (Gerevini, Haslum, Long, Saetti & Dimopoulos, 2008). Methods for generating a "diverse" set of plans (Srivastava, Kambhampati, Nguyen, Do, Gerevini & Serina, 2007) could also be used to automate complexification.

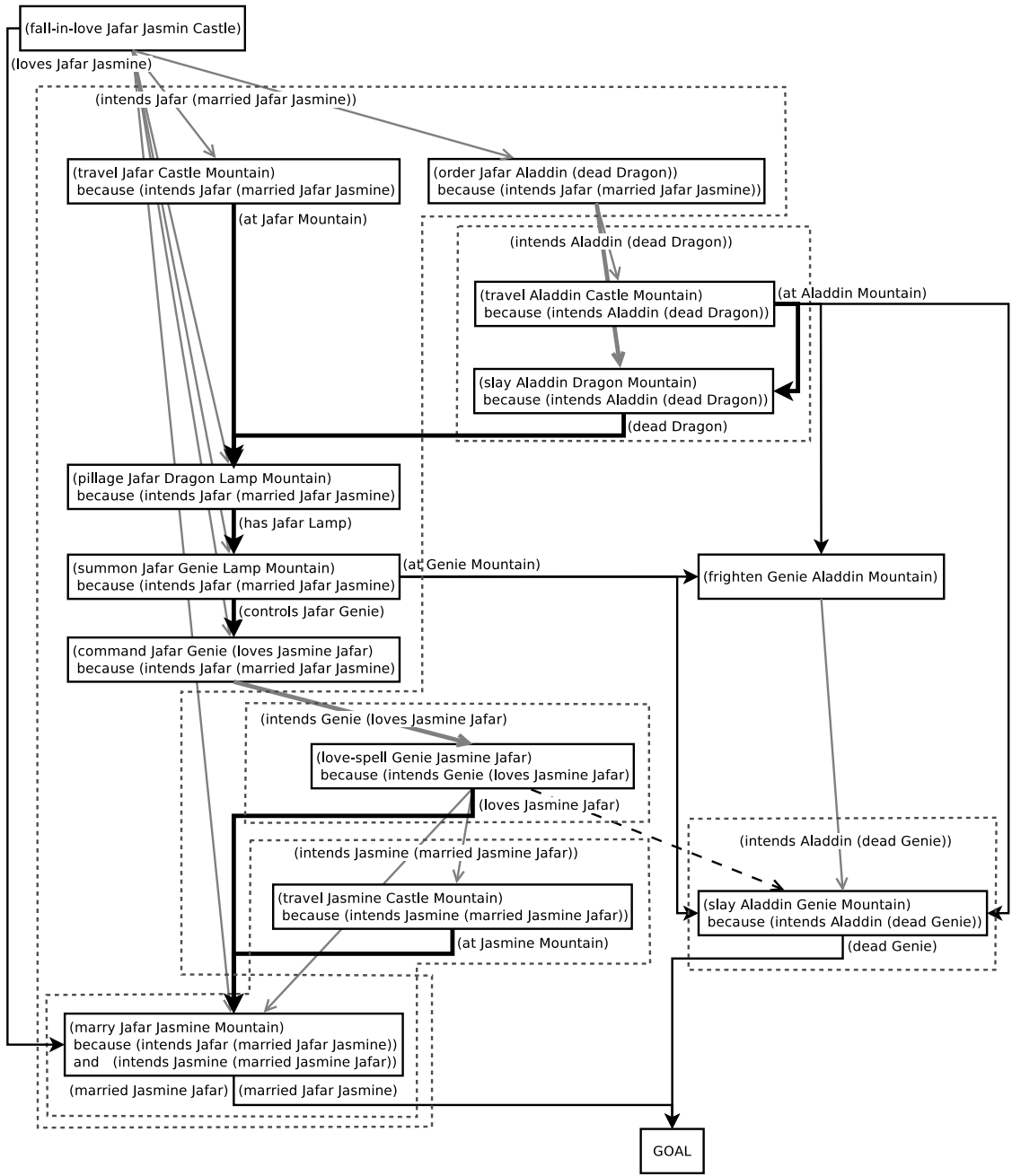

Figure 1: A story plan generated for the example problem. Motivational links are drawn in gray. The dashed edge is an ordering constraint. The outlines group actions that form a frame of commitment for a character. To avoid clutter, causal links for justified predicates are not shown; instead, causal links from the chosen effect of each action are drawn in bold. It can be seen that each chosen effect, except for final steps, links (directly or indirectly) to a precondition of an action in the same frame of commitment.

The total time to generate the plan is around 45 seconds (and of that, only half is actual search; the rest is grounding and preprocessing.) The time required for the compilation itself is less than a second. This is in stark contrast to the running time of the IPOCL planner on this problem, reported to be over 12 hours even with a problem-specific search heuristic (Riedl & Young, 2010). However, this example represents a very small problem. It contains only the actions and objects necessary to form the "intended" story plan, and no more. A more realistic scenario is a problem specification that contains many possible actions and objects that are not relevant to the story outcome, or that allow the construction of materially different plans for that goal. The size of the compiled problem can grow quite quickly as the size of the original narrative planning problem increases. As an example, a larger version of the same problem, including three more actions and a few more items, none directly relevant to achieving the outcome, takes nearly 30 minutes to solve.

## 4. Compilation 2: Meta-Planning

*"If only I had the Magic Lamp, thought Jafar. Then I could summon the Genie to gain control over it. If I controlled the Genie, I could command it to make Jasmine love me."*

The second compilation is based on simulating the characters' process of forming intentions by making plans, using explicit character planning actions. It has some similarity to Wolfe and Russell's (2011) use of explicit establishment of intentions as a means to guide plan search more efficiently. Compared to the justification-tracking compilation, it is less complex but also less stringent: plans for a meta-planning compiled problem are not guaranteed to be intentional, according to the definition of Riedl and Young, although most of the time they will be.

A meta-planning action allows a character to adopt the intention of achieving the precondition of an action that achieves a goal that the character already intends. To avoid characters making plans that they never act on, a counter tracks the number of intentions each character has, and is required to be zero at the end of the plan.[3] The counter can be represented by the standard propositional encoding (though this limits the depth of intentions a character can hold), or by a numeric fluent.

Let $(a\ \vec{x})$ be an intentional action, $x_A$ the parameter that represents its actor, and $(e\ \vec{z})$ its chosen effect. The corresponding action in the compiled problem has the additional precondition (intends $x_A$ $(e\ \vec{z})$) and effects ¬(intends $x_A$ $(e\ \vec{z})$) and decreases $x_A$'s intention count by 1. In other words, to take an action, the actor must have one of its effects in his current set of intentions, and performing the action releases the actor from that intention. If $(e\ \vec{z})$ is a modal literal of the form (intends $z_A$ $(q\ \vec{z'})$), the precondition and effect refer instead to $(q\ \vec{z'})$, and the action also increases the intention count of $z_A$, i.e., the effect is to move $(q\ \vec{z'})$ from the intention set of $x_A$ to that of character $z_A$. Happenings that add character intentions must also increase the intention count.

---

3. Some exceptions must be made: for example, if a character dies, he obviously cannot act on any outstanding intentions, but this should not invalidate the plan.

For each precondition $(p\ \vec{y})$ of $(a\ \vec{x})$, the compiled problem also has an action (plan-to-$a\ \vec{x}$), with precondition (intends $x_A\ (e\ \vec{z})$) and effects (intends $x_A\ (p\ \vec{y})$) and increasing $x_A$'s intention count. If $(a\ \vec{x})$ has several preconditions, an order of achievement among them can be enforced by adding subsets of those preconditions to the meta-planning actions. For example, the action (give ?who ?what ?to-who ?where) has the preconditions: (has ?who ?what); (at ?who ?where); and (at ?to-who ?where). Adding (has ?who ?what) to the precondition of the meta-planning action that establishes (intends ?who (at ?who ?where)) forces a character who intends to give something to not only plan to acquire the item, but to actually do so, before planning to travel (if necessary) to a place where the recipient of the gift is. Some necessary and reasonable constraints on the order of achievement of action preconditions may be found by landmark ordering analysis (Hoffmann, Porteous, & Sebastia, 2004). Manually adding further constraints to meta-planning actions gives them some flavour of (a simulation of) methods in HTN planning (Erol, Nau, & Hendler, 1994).

The reason why the meta-planning compilation does not guarantee intentional plans is that while it forces characters to motivate any action by a plan, it does not force them to monitor that their plans are still valid when the action takes place. For example, if Aladdin plans to slay the Dragon in order to pillage the Lamp, but a thief steals the Lamp from the Dragon while Aladdin is on his way to the Mountain, Aladdin still has license to slay the Dragon, even though this no longer contributes to getting him the Lamp (in fact, he must slay the Dragon to avoid being left with an unfulfilled intention). In part, this could be rectified by encoding a more elaborate structure of character plans than just the set of outstanding goals. For example, a directed graph encoding could track dependency relations between intentions, and their dependence on story world facts. This may also provide a basis for allowing characters to revise their plans in the face of changed circumstances.

Limited computational experiments with the meta-planning compilation suggest that while it produces much smaller (ground) problems than the justification-tracking compilation, these can still be harder for current heuristic search-based planners to solve.

## 5. Conclusion

Research into the classical planning problem has developed a wide array of, sometimes highly effective, methods for solving such problems. Through compilations, the capabilities of existing classical planners can be leveraged to solve many more problems than those that on the surface appear to be classical planning problems. Like the loyal knight of the story, a classical planner will committedly try to solve whatever task is set before it, as expressed by the planning domain specification. The trick is setting it the right task.

As noted, the narrative planning model defined by Riedl and Young has some limitations. For example, it does not allow to create a story in which a character tries but fails to achieve a goal. Brenner (2010) describes an approach to story generation that interleaves classical planning for individual characters' goals, based on the characters' state of knowledge, with plan "execution", i.e., adding events to the story. This permits the system to generate stories where characters are forced to abandon their plans after learning new facts, or postpone planning until crucial facts become known. Brenner claims that "it would be quite difficult to describe [such a plot] with a single plan, let alone generate it with a single planner run." It does indeed appear quite difficult, but whether or not it is impossible remains an open question.

## Acknowledgments

I wish to thank Alban Grastien, Malte Helmert, Robert Mattüller and the reviewers for useful comments on drafts of this paper. This work was supported by the Australian Research Council discovery project DP0985532 "Exploiting Structure in AI Planning". NICTA is funded by the Australian Government as represented by the Department of Broadband, Communications and the Digital Economy and the Australian Research Council through the ICT Centre of Excellence program.